\documentclass{article}
\usepackage{ijcai25}

\usepackage{times}
\usepackage{fontawesome5}
\usepackage{enumitem}
\usepackage{soul}
\usepackage{url}
\usepackage[hidelinks]{hyperref}
\usepackage[utf8]{inputenc}
\usepackage[small]{caption}
\usepackage{graphicx}
\usepackage{amsmath}
\usepackage{amsthm}
\usepackage{booktabs}
\usepackage{algorithm}
\usepackage{algorithmic}
\usepackage[switch]{lineno}

\usepackage{siunitx}
\usepackage{natbib}

\def\o1{o1}

\title{ Evaluating the Ability of Large Language Models\\
to Reason about Cardinal Directions, Revisited\footnote{This paper is an update on a paper published at COSIT-24 \citep{cohn_et_al:LIPIcs.COSIT.2024.28}, but we have conducted many further experiments since then as the number and variety of LLMs has continued to multiply.}}

\author{
Anthony G Cohn$^{1,2}$
\and
Robert E Blackwell$^{2,1}$
\affiliations
$^1$School of Computer Science, University of Leeds, UK\\
$^2$Alan Turing Institute, London, UK\\
\emails
a.g.cohn@leeds.ac.uk,
rblackwell@turing.ac.uk
}
\begin{document}

\maketitle

\begin{abstract}
We investigate the abilities of 28 Large language Models (LLMs) to reason about cardinal directions (CDs) using a benchmark  generated from a set of templates, extensively testing an LLM's ability to  determine the correct CD given a particular scenario.  The templates allow for a number of degrees of variation such as means of locomotion of the agent involved, and whether set in the first, second or third person. Even the newer Large Reasoning Models are unable to reliably determine the correct CD for all questions. This paper summarises and extends earlier work presented at COSIT-24.
\end{abstract}

% In 2024 we released a benchmark to investigate the abilities of a representative set of Large language Models (LLMs) to reason about cardinal directions (CDs) \citep{cohn_et_al:LIPIcs.COSIT.2024.28}. Much has changed in the intervening period including the emergence of Large Reasoning Models (LRMs).

\section{Introduction}
This paper summarises and extends an earlier paper which investigated the ability of Large Language Models (LLMs) to reason about cardinal directions \citep{cohn_et_al:LIPIcs.COSIT.2024.28}.  That paper introduced two benchmarks: a small dataset of 100 questions designed to test simple factual understanding of cardinal directions, and a more complex one, with 5760 questions, generated from a set of six templates which requires some spatial reasoning to answer correctly. Already in the earlier paper several LLMs were obtaining very good performance on the small dataset, and newer models do even better.   Therefore this update concentrates solely on the larger dataset and documents the performance of newer LLMs on it,  conducting a more thorough analysis overall.

Many claims (e.g. \citep{creswell2022faithful,huang2023reasoning,kojima2022large}) have been made since the emergence of Large Language Models (LLMs) as to their ability to reason. Spatial reasoning is of particular interest since not only does it underlie a  human's  ability to operate in the physical world, but also because LLMs are not embodied; %and thus have no actual experience of the physical world;
so the question arises, have they nonetheless acquired an ability to reason about situations which might occur in the real world?  This is the question we address here. Spatial reasoning in general encompasses many aspects including topology, distance, and direction.  Here, we restrict ourselves to  reasoning about
%directions, and in particular
cardinal directions (CDs). CDs are important for many reasons, e.g.: (i) successful navigation and route finding/following usually requires a fundamental understanding and ability to reason about CDs: CDs are crucial to success when using a map. Equally, giving directions often relies, at least in part, on the use of CDs. (ii) Understanding the geography of
an area
%a local area, or
%at larger scales
%indeed at a national, continental or intercontinental scale
depends on understanding the relative CD of one location to another -- migration patterns, climate differences and economic variations are often underpinned by CDs. (iii) Weather patterns are often heavily influenced by the direction of the prevailing wind. (iv) CDs often play a critical role in cultural and historical contexts, e.g. the alignment of the pyramids to the four
%main
CDs, or in certain languages -- e.g. the aboriginal language \emph{Guugu Yimithirr}  has no words for left or right, and  spatial information is mainly conveyed using CDs \citep{haviland1998guugu}.

In this paper we therefore analyse how well LLMs
%(e.g. GPT-3.5-turbo, GPT-4,..)
can reason about cardinal and intercardinal directions.
%in spatial contexts.
We do this by automatically constructing a large set of
%benchmark
questions based on templates, for which the correct answer has been pre-determined, and testing each LLM's ability to answer the questions correctly.

In the original paper, Large Reasoning Models (LRMs), in which the models have been trained to perform inference-time reasoning,  had not yet emerged, though chain-of-thought prompting (CoT) \citep{wei2022chain}  was already popular\footnote{We  did not explore CoT as we wanted to measure raw LLM performance, independent of any particular prompting strategy; we do not explore CoT prompting in this paper either.}.  In the intervening months, LRMs have generally become  the most performant LLMs, and this paper provides an update on how LRMs perform on our dataset, as well as some other new LLMs, such as GPT-4.5 preview.  Since many LLMs now claim to have reasoning capability, we use the strict definition that an LRM is one which reports the use of reasoning tokens
in its response to
prompts in our benchmark. In this paper we investigate the following LRMs:
azure-o1-2024-12-17, azure-o3-mini-2025-01-31,
azure-o3-mini-2025-01-31-high, azure-o4-mini-2025-04-16, and
deepseek-reasoner. For futher information on LRMs see  \citet{xu2025largereasoningmodelssurvey} who provide a survey covering the emergence and architectures of LRMs.

% We also tested the LLMs on a small set of simpler questions, co-created with ChatGPT, in which recall of world knowledge is more prevalent.

\section{Related Work}
%We refer the reader to the earlier paper \citep{cohn_et_al:LIPIcs.COSIT.2024.28} for a more extensive literature review; here we focus on work that has emerged since that paper went to press.
Despite the rapidly growing amount of research into LLMs and their capabilities there has been relatively little devoted specifically to spatial and/or geographic reasoning, and none which has  tested their ability to reason about CDs in the way we do here. Of the existing work, we note benchmarks such as StepGame \citep{li2024advancing,shi2022stepgame} which aims to test an LLM's ability to correctly determine the spatial relationship between two objects, given the spatial relations between a larger set of objects, and between 1 and 10 reasoning steps are required to correctly determine the result;  the direction relationships are not exclusively  CDs, but also include, for example ``clock face directions'' (B is in the three o'clock direction from C). Not surprisingly, performance deteriorates as the required number of steps increases. However, performance increases markedly when an LLM is used to translate from the English specification to a logical representation and symbolic reasoning is used to compute the relationship. It should also be noted that the setting in StepGame is rather artificial compared to the naturalistic settings in our dataset described below. The SpartQA dataset \citep{mirzaee-etal-2021-spartqa} is also focused on assessing spatial reasoning, but does not contain any CDs.  The bAbI dataset \citep{weston2016towards} has one task which tests CDs understanding,  task 19,  which contains 1000 training and 1000 test questions: each instance contains 5 facts stating CDs between two objects, and then a question asking about the relation between two of them. Other work \citep{yamada2024evaluating} has investigated whether LLMs can acquire an understanding of a spatial environment from a turn-by-turn description of a route, with landmarks named at each turn; whilst the LLMs did perform reasonably well, the experiment did not involve any CDs, only left/right and up/down.

There are different kinds of spatial reasoning tasks which can be considered.  Relational composition is one of the most studied from a theoretical point of view.
%(given a relation R1 between $x$ and $y$, and a relation R2 between $y$ and $z$, what is the relation between $x$ and $z$?).
A \emph{composition table} records the results for all combinations of relations in a particular spatial representation such as RCC;  investigations \citep{cohn2023evaluation,cohn2024largelanguagemodelsreason} into various LLMs' abilities to compute all RCC compositions found reasonable accuracy levels (reduced when  relations are anonymised); however RCC is a purely mereotopological calculus with no notion of direction embedded in its semantics.

Some LLMs have been built specifically for geo-applications, but these do not focus on reasoning about directions but rather aspects such as toponym recognition, e.g. \citet{Li2023GeoLMEL}.

\section{Experimental Design}
Whilst  testing compositional reasoning with CDs would be of interest, here we restrict ourselves to testing simpler reasoning abilities.
% We created two question and answer sets which we refer to as \smallq{} and \largeq{}.
% %
% For \smallq{}
% %the small question and answer set,
% we used ChatGPT to co-create 100 simple questions where the answer is a CD \( \{north, south, east, west\} \). We edited the questions and corrected the answers where necessary. We changed the questions to ensure equal class representation amongst the four answers. Example questions are:
% %real world
% %questions requiring world knowledge and simple reasoning, e.g.:
% \begin{itemize}
% \item \textit{You are watching the sun set. Which direction are you facing?}
% \item \textit{If the South Pole is behind you, which direction are you facing?}
% \end{itemize}
% We use this dataset to give an overall assessment of LLM performance in real world scenarios requiring directional common sense spatial reasoning and common sense spatial knowledge.
%
It would be impractical to generate a substantial question set manually and so we used an automated, template driven approach.
%for
%our second (large) question set.
%\largeq{}.
We wanted to investigate the ability of LLMs to reason about CDs in the context of a simple scenario involving locomotion along or around a geographical feature as a test of an LLM in a realistic situation. Based on some informal experimentation using GPT and ChatGPT with a selection of questions and noting a lack of accuracy,   six question templates (T1-T6, shown in Figure \ref{fig:accuracy_aspects}) were chosen to test LLM performance more comprehensively. Note that five of these concern areal objects (lake, island, park), one concerns a linear object (a road, T4); of the areal objects, the island differs from the other two, in that when on the, say, east shore of an island the sea is to one's east, differently to being on the east shore of lake, where the lake is to the west.
% \begin{itemize}[nosep]

% \item[T1] \textit{You are walking [south] along the [east] shore of a lake; in which direction is the lake?}
% \item[T2] \textit{You are walking [south] along the [east] shore of a lake and then turn around to head back in the direction you came from, in which direction is the lake?}
% \item[T3]  \textit{You are walking [south] along the middle of the [east] side of a park; in which direction is the bandstand located in the centre of the park?}
% \item[T4]  \textit{You are walking [east] along the [south] side of a road which runs [east to west]. In which direction is the road?}
% \item[T5]  \textit{You are walking [south] along the [east] shore of the island. In which direction is the sea?}
% \item[T6]  \textit{You are walking [south] along the [east] shore of an island and then turn around to head back in the direction you came from, in which direction is the sea?}
% \end{itemize}

We then exhaustively generated all forms of these questions for all cardinal and intercardinal directions and ten different locomotion types
\emph{\{cycling,
driving,
hiking,
jogging,
perambulating,
racing,
riding,
running,
unicycling,
walking\}}. Note that in each template, once one of the directions (between ``[ ]'') is fixed, then there are only two possibilities for the second direction\footnote{Note that for T4, there could potentially be twice the number of choices if one were to consider, e.g.  ``[west-east]'' as well as ``[east-west]''.  This could conceivably make a difference to an LLM since in one case the agent is walking in the direction specified, and in the other case in the opposite direction.}.  Following earlier evidence \citep{raman2024steer} that an LLM's performance can vary depending on whether  a question is phrased as using the first or second person, we also generated questions in the first-person (\emph{I am}), first-person plural (\emph{We are}), second-person (\emph{You are}), third-person singular (\emph{He is} and \emph{She is}), and third-person plural (\emph{They are}) forms. This gives us 6 templates $\times$ 10 locomotion forms  $\times$ 6 person forms $\times$ 8 directions $\times$ 2 directions-variations = 5760 questions.

       % ;; Example 4 - You are walking [east] along the [south] side of a road
       % ;; which runs [east to west]. In which direction is the road?

       % (expand
       %  (:question (format nil "~a ~a ~a ~a along the ~a side of a road which runs ~a to ~a. In which direction is the road?"
       %                     (noun person) (verb-to-be person) locomotion (rotate direction 90) direction (rotate direction 90) (rotate direction 270))
       %             :template "t4"
       %             :absolute-answer (rotate direction 180)))

       % (expand
       %  (:question (format nil "~a ~a ~a ~a along the ~a side of a road which runs ~a to ~a. In which direction is the road?"
       %                     (noun person) (verb-to-be person) locomotion (rotate direction 270) direction (rotate direction 270) (rotate direction 90))
       %             :template "t4"
       %             :absolute-answer (rotate direction 180)))

Previous work suggests that models with less than about 40B parameters perform poorly at reasoning \citep{raman2024steer}.
We test state-of-the-art models, favouring larger
%commercial
offerings.
However, we did test four smaller models (OLMO 2, DeepSeek R1 7B, LLama 3 7B, and Mistral Small 24B) which were able to run locally on an Apple M2 Pro MacBook using Ollama\footnote{\url{https://ollama.com}, accessed May 2025.} for comparison.
All LLM experiments were conducted using the Golem software\footnote{\url{https://github.com/RobBlackwell/golem}, accessed May 2025.}.
An advantage of Golem is that all HTTP API requests and responses are logged, providing a comprehensive record of
experiments and results in JSONL\footnote{\url{https://jsonlines.org}, accessed May 2025.} format. Unlike some other benchmark studies, all model versions, settings, hyper-parameters, and results are preserved for subsequent analyses, facilitating transparent and reproducible science, as advocated by \citet{burnell2023rethink}.

%We therefore favour larger models, testing those listed in Table~\ref{tab:models}.

% \begin{table*}[htb]
% \footnotesize
% \centering
% %\begin{tabular}{|m{2.5cm}|m{4cm}|m{1.5cm}|m{1cm}|m{1.5cm}|}
% \begin{tabular}{|c|c|c|c|c|}
% \hline
% \textbf{API} & \textbf{Model} & \textbf{Released} & \textbf{
% %Size
% Num. params} & \textbf{Window} \\ \hline
% Anthropic Claude& claude-3-opus-20240229 & Feb 2024 & 137B & 200,000  \\
% Google Vertex & gemini-10-pro & Dec 2023 & 1.6T &  32,000   \\
%  & gemini-15-pro-preview-0409 & Apr 2024 & $\gg$3.5T & 128,000   \\
% OpenAI & gpt-3.5-turbo-0613 \cite{brown2020language}  & Jun 2023 & 175B & 4,096   \\
%   & gpt-35-turbo-1106 & Nov 2023 & 175B & 16,385  \\
%   & gpt-3.5-turbo-0125 & Jan 2024 & 175B & 16,385  \\
% & gpt-4-0613 \cite{openai2024gpt4} & Jun 2023 & 1.76T & 8,192   \\
%   & gpt-4-turbo-2024-04-09 & Apr 2024 & 1.76T & 128,000   \\ \hline
% \end{tabular}
% \caption{LLMs tested.
% in our experiments.
% %Size is the number of parameters.
% Window is the context window size in tokens.}
% \label{tab:models}
% \end{table*}

\subsection{Prompting}

%Zero-shot prompting is when a model is given a question without any
%prior specific training.
%examples to help guide the answer.
%The model must then attempt to answer the question based solely on its general pre-training.
We used zero-shot prompting with a system prompt of ``You are a helpful assistant. I will give you a question about directions. The answer is either north, south, east, west, north-east, north-west, south-east or south-west. Please only reply with the answer. No yapping.". We then presented each question in a new chat.  We included ``No yapping." since that has been reported as being beneficial in persuading an LLM to be brief in its response \citep{stifter2024unit}.

Where models allow the temperature to be explicitly set (all models except \o1, o3-mini, o4-mini and GPT4.5),
we set temperature = 0 to try to achieve deterministic answers, although we show \citep{blackwell2024reproduciblellmevaluationquantifying} (and in the results below) that this cannot in general be achieved on models run in the cloud.

% We set temperature = 0 to try to achieve deterministic answers for those  models (marked with {\tiny \faThermometerHalf}\  in Fig. \ref{fig:summary}) which allow this.

%Temperature is a parameter that affects the randomness or variability of the responses generated by a language model and helps to control the predictability of the the model's output but even a 0 temperature does not guarantee reproducible, deterministic behaviour.
%
%\rob{Not all models (especially LRMs) allow us to change temperature.}
% To explore the effect of temperature on accuracy, we take our best performing model on the \emph{large} dataset and vary temperature \(t\) , \(0 \leq t \leq 2\).
%
We use case-insensitive string comparison and remove spurious punctuation and white space before comparing answers.
%; e.g. we regard ```North East'." and ``north-east" as equivalent.
Our prompts ask for cardinal or intercardinal direction answers only,  but allow answers such as ``The lake is to the west"  if the intended answer is ``west",  recording such instances where answers do not strictly meet the rubric. We assess performance using accuracy, using
prediction intervals \citep{blackwell2024reproduciblellmevaluationquantifying} to report variability (where practical).
%the standard error of the mean.

\section{Results}

\begin{figure}[ht!]
    \centering
    \includegraphics[page=1, width=\columnwidth]{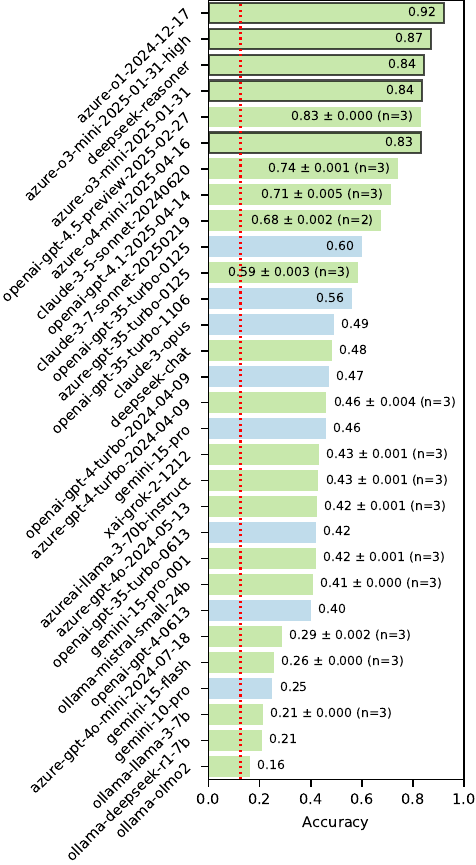}
    \caption{
    Accuracy by LLM. Results shaded in blue are from \citet{cohn_et_al:LIPIcs.COSIT.2024.28}. Results with a black border are LRMs. The red dotted line is the guess rate (0.125, since there are eight possible answers). Where possible we use the model names and versions reported in the LLM response, prefixed by the API provider (e.g. openai or azure). Models run locally are prefixed with ollama.}
    \label{fig:summary}
\end{figure}

None of the models tested are accurate for all questions (Fig. \ref{fig:summary}), although all models have higher accuracy than the guess rate.
The best performing model in \citet{cohn_et_al:LIPIcs.COSIT.2024.28} was GPT3.5 Turbo (0.60), but the best performing model now is \o1 (0.92), an increase in
accuracy of more than 50\% in a year.
The top 22 results are all commercially-hosted Large Reasoning Models.
The best performing locally-hosted model is mistral-small-24b (0.41).
%
% See discussion section for cost calculations
%
In the original paper we only ran models once, and have not conducted further runs here;   new models were run more than once to obtain error bars (even with temperature = 0 there can be non determinism \citep{blackwell2024reproduciblellmevaluationquantifying}), except when cost precluded this  -- many LRMs are expensive, e.g. \o1 cost
\$458 for one run.

We also see  slightly different accuracy when accessing the same OpenAI model via the Azure API versus OpenAI API (GPT3.5-Turbo and GPT4-Turbo) and
further analysis suggests that these are statistically different \citep{blackwell2024reproduciblellmevaluationquantifying}.

\begin{figure}[ht!]
    \centering
    \includegraphics[page=1, width=\columnwidth]{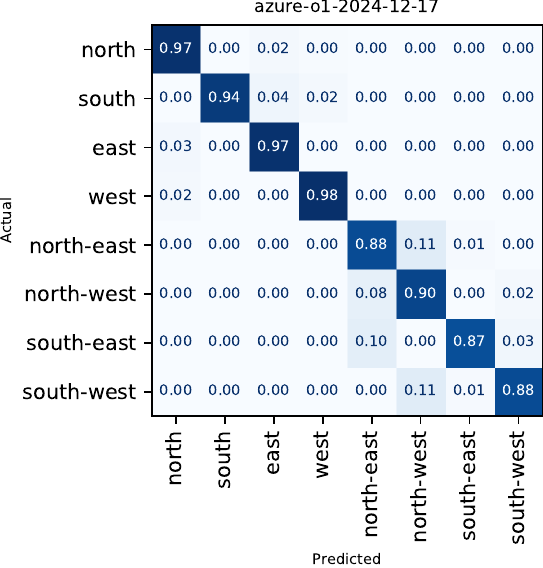}
    \caption{
    Confusion matrix for the best performing model, \o1.}
    \label{fig:cm}
\end{figure}

Where confusion occurred, it was mostly with intercardinal directions rather than cardinal directions (e.g. Fig. \ref{fig:cm}).
It can be seen that confusions are not symmetric in general which is perhaps revealing about the reasoning process, as a purely logical reasoner would be expected to produce symmetric computations.

% All models tested showed an accuracy of $> 0.8$ for \smallq{}  (Fig.~\ref{fig:summary}a). Where confusion occurred, it was mostly \textit{north} confused with \textit{south}, and \textit{east} confused with \textit{west} (Fig.~\ref{fig:summary}c). In one case a model ignored the rubric: gemini-1.0-pro  answered the question \textit{"On a hike, a duck pond is to your north and the nearest town is south. What direction is the pond from the town?"} with \textit{"The pond is north of the town"}, which is correct but not a one word answer. Of the 100 questions, 77 were always correctly answered.
% Only one question was always answered incorrectly:
% \textit{In a stadium with a north-facing entrance, if the VIP section is on the left side, which direction would it be in? (east)}; all answered \textit{west}.

\begin{figure}[ht!]
    \centering
    \includegraphics[page=1, width=\columnwidth]{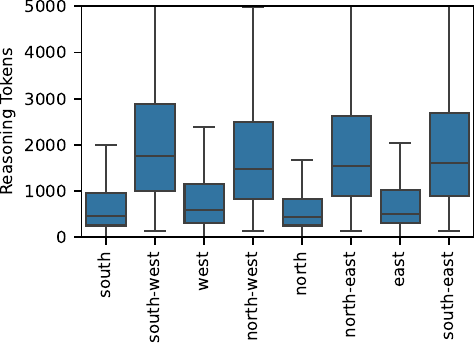}
    \caption{
    Median and inter-quartile range of \o1 reasoning tokens by ground truth direction.   }
    \label{fig:tokens}
\end{figure}

{\bf Reasoning tokens consumed}:  LRMs report the number of completion tokens used for reasoning about a question, and we investigate the distribution
of the count of reasoning tokens by correct direction (See Fig. \ref{fig:tokens} for \o1). More completion tokens are used for questions involving
intercardinal directions than cardinal directions. We see a similar pattern for the other LRMs tested. We also note that the interquartile range is much larger for the intercardinal directions. Finally, we observe that, for \o1, north is the easiest cardinal direction and east the most difficult, whilst north-west appears to be the easiest intercardinal direction, and southwest the most difficult.

\begin{figure}[ht!]
    \centering
    \includegraphics[page=1, width=0.98\columnwidth]{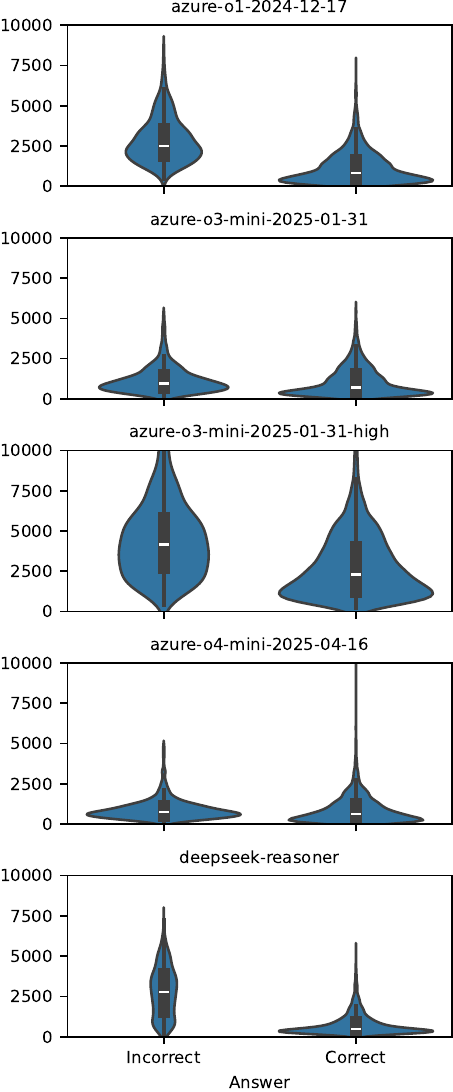}
    \caption{
    Reasoning token counts for correct and incorrect answers by LRM.
    The white bar shows the median.
    Note that all LRMs tested have more correct than incorrect answers
    and so incorrect sample sizes are small.}
    \label{fig:reasoning}
\end{figure}

We also investigated how many reasoning tokens were consumed when the LRMs got the answer right compared to when they got the answer wrong (see (Figure \ref{fig:reasoning}). Three of the LRMs (o1, o3-mini High, and Deepseek R1) have a noticeably higher median count of reasoning tokens used for incorrect answers
than correct answers.
o3-mini uses more reasoning tokens when reasoning\_effort\footnote{\url{https://platform.openai.com/docs/guides/reasoning?api-mode=chat}} is set to ``High" rather than the default ``Medium". o4-mini has a slightly lower median for correct compared to incorrect answers but has a very long  but small tail of correct answers requiring a very large number of reasoning tokens.

\begin{figure*}[ht!]
    \centering
    \includegraphics[page=1, height=0.98\textheight, width=0.98\textwidth, keepaspectratio]{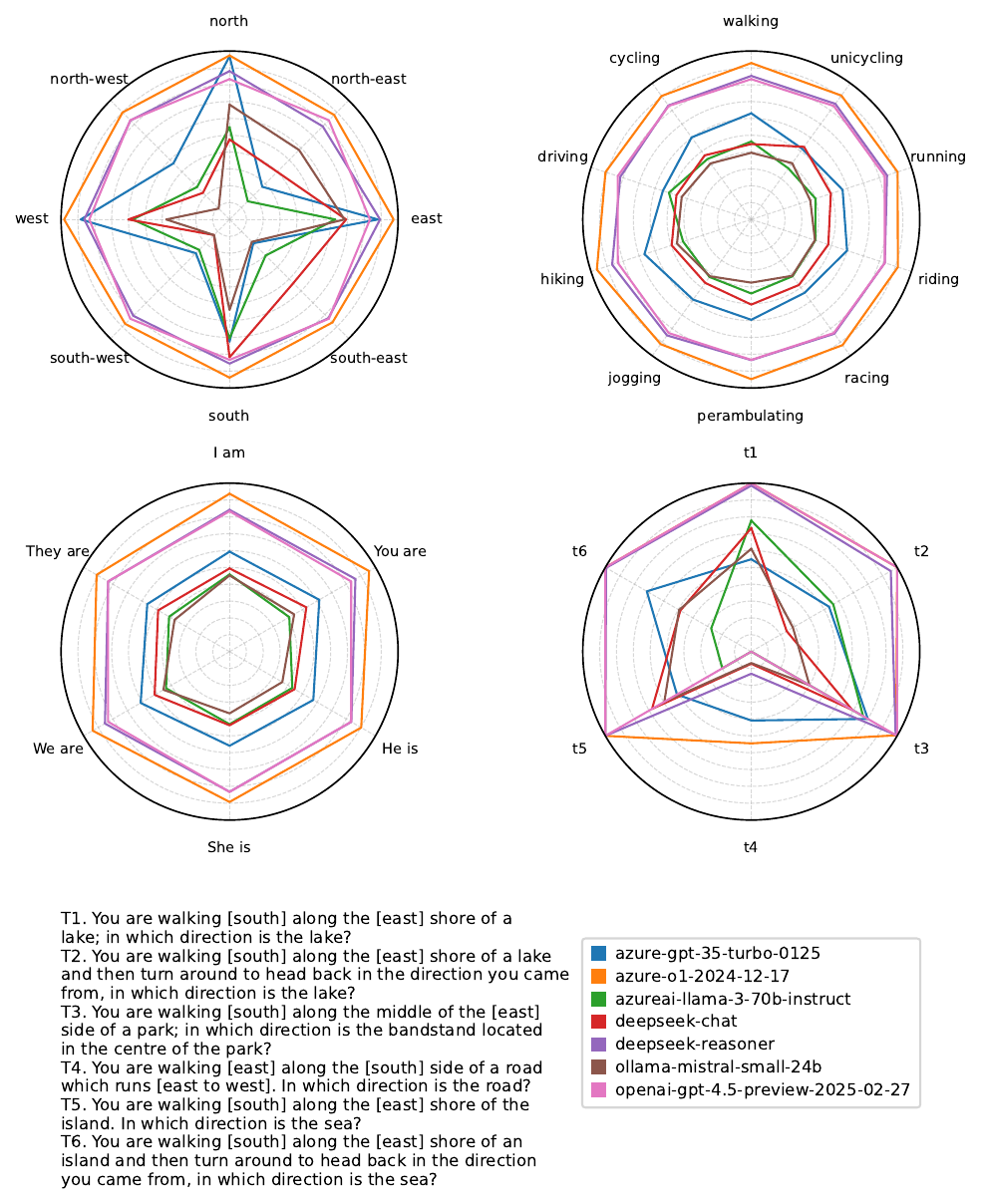}
    \caption{Accuracy by direction, locomotion, person form, and question template for selected models. The grey concentric circles in the background are set at an accuracy of 0.1 apart.}
    \label{fig:accuracy_aspects}
\end{figure*}

Fig. \ref{fig:accuracy_aspects} analyses seven models' performance on CDs,  person,  locomotion and template -- we choose just seven models since looking at all the models in Fig. \ref{fig:summary} would be overwhelming. We chose the best model overall (\o1), the best non LRM (GTP4.5 preview), the best model in the earlier paper (GPT3.5T), the  best LRM not from OpenAI (Deepseek-reasoner), the deepseek model from which this was built (Deepseek-chat), the largest open weights model tested (LLama 3 70B instruct) and the best model which can be run on a laptop (mistral-small-24b). The key point about varying the person form and the locomotion type is that such changes should not make any difference to the accuracy as the result is clearly (at least to a human) independent of such variations; but as can be seen in the top right and lower left charts in Fig. \ref{fig:accuracy_aspects}, there can be considerable variations in accuracy.  For cardinal directions,  one would have expected that if a model can solve for one choice of directions in the prompt, then swapping to a different set is mathematically/formally equivalent, but again we see this is not the case.

{\bf Cardinal Directions}: all models except GPT4.5\footnote{Please see our conclusions about GPT4.5 which we suspect may have been trained on our benchmark.} had higher overall accuracy with cardinal than intercardinal directions\footnote{ It is interesting to note that humans have been reported to be worse at ``oblique" directions, rather than the two orthogonal axes \citep{,MENG20053402,appelle1972perception} -- the so called ``oblique effect''. It is suggested to have a neuropsychological basis, though we are not suggesting the effect here is for the same reason. A more likely explanation, in the world of large {\emph{language}}  models, might be that the intercardinal relation names are composed of two words (in English and many other, but not all languages), and this might create confusion in an LLM.  } though there were some notably strange results such as with Claude Sonnet where for example, South East had higher accuracy than North (Claude Sonnet 3.5: 0.825 / 0.614, Claude Sonnet 3.7: 0.801 / 0.607). The previously best model, GPT-3.5T still performs very competitively on the cardinal directions but  is let down on its performance on the intercardinals, suggesting that the reasoning available in LRMs really helps for intercardinals (though GPT-4.5 which is not a reasoning model has similar performance across both CDs and intercardinals).  Deepseek-reasoner and GPT-4.5  have a very similar performance, though with GPT-4.5 slightly outperforming Deepseek-reasoner on the intercardinals, whilst the situation is reversed on the cardinal directions.  It is striking how much better Deepseek-reasoner is compared to the base model, Deepseek-chat, from which it was trained. The smallest model, Mistral, which was run on a laptop, is the bottom of the pack, except for north, northeast, and east, yielding an unexplained asymmetry.

{\bf Person form}: It can be easily observed that the top five models form  ``concentric hexagons'', showing reducing performance from  \o1 through to Deepseek-chat, with roughly similar performance for each person-case;  the bottom two LLMs,  Llama-3-70b and Mistral-24B  jostle for position with Llama outperforming Mistral on ``He is",  ``They are", and  ``She is", but slightly worse on ``You are".  GPT-3.5T is worse at ``she is" versus ``he is" suggesting a possible slight gender bias (possibly as a result of biased training data)  -- this is also true of some of the other models not selected for this figure but in Fig. \ref{fig:summary}.

{\bf Locomotion form}: The best four models  along with Llama-3-70b  form concentric decagons, showing strictly reducing accuracies. However  Deepseek-chat and Mistral-small-24b show more variation with each outperforming their nearest rival for unicycling, and the latter outperforming Llama-3-70b for hiking. \o1 does best overall on hiking, which maybe surprising when according to our investigations walking is likely to appear more frequently in the training data -- but perhaps hiking appears more frequently in contexts where cardinal directions are used (such as trail maps/instructions).  The original model, GPT-3.5T is best for walking and hiking, and worst for unicycling, racing and driving.

% \tony{Yes, I see what you mean. I guess the locomotion, direction and person ones are slightly more interesting, but with the error bars, there's probably no statistically significant difference between the different bars?  Perhaps we should report on one of the not quite so good models? (as well/instead of?).  Also the tokens usage for correct vs incorrect might be interesting to report -- more tokens to get an incorrect answer?}

{\bf Template variation}: If we look at the performance of all the seven selected models  across the six different templates  (see Fig. \ref{fig:accuracy_aspects}, bottom right) then the most striking aspect is that  T4 (the only template concerning a linear rather than an areal object) has the worst accuracy, by a substantial margin (this is also the case for the other LLMs in Fig. \ref{fig:summary}. We do acknowledge that the wording of T4 could be improved but this alone would not seem to account for this effect.  Worst of these was GPT-4.5 preview  which got all of the T4 questions wrong (91\% of the time it gives an answer 90 degrees off (i.e. consistent with the road direction). The remainder of the time it gives an invalid answer echoing the road description, e.g. north-south) -- and all the other questions on the other templates completely correct; given the probabilistic nature of LLMs this seems statistically unlikely, but we do not have a good explanation for it.

\o1 does best on T4 (0.55), followed by GPT-3.5T (0.41); apart from its performance on T4, GPT-4.5 is pretty similar to \o1. T2 is a slightly more complicated variant of T1 (since it involves ``head[ing] back in the direction you came from'' and it can be seen that this results in a slight performance dip for each of the models. T3 is very analogous to T1, except for a change of scenario from lake to a park; many models achieve similar accuracy, though Deepseek-chat is worse, as is Mistral-small-24b, whilst GPT-3.5T has better accuracy. T5 and T6 are similar to T1 and T2 respectively, but the answer is the opposite in in each case (a lake is west of its east shore but the sea is east of the east shore of an island); \o1 and GPT-4.5 have very similar performances in each case, whilst GPT-3.5T is about the same on T1/T5 but better on T6 compared to T2. We might have expected LLMs to confuse the direction of the lake (T1,T2) and the sea (T5,T6), and 180 degrees is indeed the most common confusion for five models (o3-mini, o4-mini, Claude 3.5 Sonnet, Claude 3.7 sonnet, and gemini-15-pro). However in most cases (11 models) the most common confusion is 90 degrees from the correct answer suggesting that confusion is not systematic.

\section{Discussion and Conclusions}

None of the models tested is able to reason completely reliably  about cardinal direction; however all models show some capacity for spatial reasoning, and \o1 in particular.  A model that randomly selects cardinal or intercardinal direction answers to an MCQ would have an accuracy of 0.125, but even the worst performing model
%(gemini-10-pro)
OLMO2
achieved an accuracy of
%0.25
0.16.

None of the models is able to fully generalise person form or locomotion. If a model was truly reasoning we would not expect the
mode of locomotion to have any impact on accuracy, nor would the gender of the agent.

Differences in results amongst models of the same name but with different versions and different vendors (e.g. GPT3.5T versions 0613, 1106, and 0125 hosted on OpenAI and Azure) highlight the importance of recording precise experimental conditions in benchmark studies.

As noted, OpenAI GPT4.5 preview (launched 27 Feb 2025) gets all questions correct except for Template 4 questions, where it gets those all incorrect. These results are suspicious and we wonder whether GPT4.5 has been trained on our benchmark.

% 0.42 to 0.6 is a 43\% increase

The development of LLMs is progressing rapidly (though many believe they will never achieve AGI, let alone ever achieve reliable reasoning abilities, at least without a neuro-symbolic component\footnote{\citet{li2024advancing} shows that the StepGame benchmark can be almost perfectly solved by using an LLM to translate the problem instances into logical form and then using an ASP reasoner to actually solve the problem.}).

The early success of prompting strategies such as Chain-of-Thought (CoT) and Tree-of-Thought (ToT) have led to
models that have been explicitly trained for reasoning at inference time. These so-called
Large Reasoning models such OpenAI’s \o1 and DeepSeek’s R1 comprise our four best performing models.

An analysis of reasoning tokens use by \o1 shows that more tokens are required for intercardinal rather than cardinal directions.
We posit that completion tokens are a proxy for reasoning effort and that intercardinal directions require more reasoning effort, since they are defined in terms of    two cardinal directions.

% The development of LLMs is progressing rapidly (though many believe they will never achieve AGI, let alone ever achieve reliable reasoning abilities, at least without a neuro-symbolic component):
% %(though there is some evidence that performance improvement is plateauing~\cite{plateauing}).
% the Open AI GPT-35-turbo model was updated twice in seven months. Using  \largeq{}  as a benchmark, we observed a 43\% increase in performance between the Open AI GPT-35-turbo 0613 and 0125 versions. However, any evaluation such as this can only ever be a snapshot evaluation, so we hesitate to draw conclusions in general as to which LLM (family) is better than another.

% We also tested  Microsoft Azure API access to  gpt-35-turbo and gpt-4. Although the Microsoft Azure API is designed to include additional guardrails, compliance and data governance certification and enterprise support, we found that accuracy was similar to the Open AI API models gpt-35-0613 and gpt-4-0613 respectively. The Azure documentation does not specify which OpenAI model their models exploit.

% GPT-4.5 cost calculation
% 1533312 prompt tokens * \$75 per million = \$115, 43730 completion tokens  * \$150 per million = \$6.55
% This is a surprsingly low number of tokens, probably caused by no yapping

% o1 calculation
% 505344 prompt tokens * \$15 per million = \$7.58, 7510046 completion tokens * \$60 per million = \$450$
% o1 pro is ten times more expensive
% $75.8 + $4500

The latest and most sophisticated large reasoning models are expensive to test. Our \o1 experiment
cost approx. \$458 to run with no repeats. By contrast, the OpenAI model, \o1 pro\footnote{\url{platform.openai.com/docs/models/o1-pro}, accessed May 2025} costs \$150 for 1M input
tokens and \$600 for 1M output tokens, putting it beyond our budget as it would have cost $\sim$\$4.6k to run a single repeat!

Some LRMs take much longer to respond to prompts than traditional LLMs. In our experiments, Deepseek R1 took a median of 34s to reply (\o1 7s, GPT4.5 0.9s, GPT 3.5 Turbo 0.7s). Our 5760 benchmark question set takes more than 48 hours to complete sequentially using Deepseek R1.

%OpenAI GPT4.5 preview (launched 27 Feb 2025)
% Not claimed to be a reasoning model?
% Suspiciously, it gets all COSIT large questions correct except for Template 4 questions, where it gets those all incorrect!
% Hence accuracy = 0.83 (5/6)
% 5th place in the ranking
% Game over for this benchmark?
% Has GPT4.5 been trained on our benchmark???

Although we have not presented further results on newer LLMs on the small dataset in \citet{cohn_et_al:LIPIcs.COSIT.2024.28} it is worth noting that we did run all the models reported here on the small dataset and most models had an accuracy of 0.9 or greater. One point of interest is that in one case\footnote{``On a hike, a duck pond is to your north and the nearest town is south. What direction is the pond from the town?”}, Ollama deepseek-r1:7b (a distilled version of Deepseek's best reasoning model, R1), we found that with a temperature of 0 it did not terminate (or at least we cut it off after 3 hours on a single prompt);  however, when given the same prompt and with the default temperature it terminated very quickly and gave the correct answer.

%\section{Future work}

%We aim to consider turn-based navigation.

%We want to consider few shot prompting

%Use of CoT and follow up questions to explain reasonoing. Explainability or interpretability is the ability of a model to
%explain or to present its predictions in understandable terms to a
%human. \cite{luo2021local}

Possibilities for future work include:
\begin{enumerate}[nosep]
\item Improving the question design; there are minor flaws in our current questions, e.g. differing punctuation in T1-T6, and a potential ambiguity in T4 (since a road is a linear object it might have been clearer to add ``from $\langle agent \rangle$" to make it clear that the question is not relating to the orientation of the road.
\item Exploration and/or  development of prompting strategies~\citep{bhandari2024survey} to improve performance -- either using general methods such as \emph{chain of thought} or \emph{tree of thoughts}, or spatial-specific ones such  as \emph{Visualization-of-Thought} \citep{wu2024visualization}.
\item Other LLMs could be evaluated, or existing ones fine-tuned for the tasks under consideration.
\item Extend the variety of experiments to create a comprehensive benchmark for evaluating reasoning about directions -- in this paper we have deliberately only considered questions whose answer is a CD, but a more comprehensive dataset would also consider egocentric directions (left, right, behind, above...) with respect to an actor or other objects with an intrinsic front,  or the landscape, or architectural artefacts.
\item We noted earlier that there are languages where the intercardinal relation names are not composed of pairs of the cardinal names, for example: Breton, Finnish and Estonian (https://en.wikipedia.org/wiki/Cardinal\_direction), and it would be interesting to see if LLMs do not have this effect of reduced accuracy for the intercardinals in such languages. Experimenting with other languages would also be of interest more generally too.
\item Building a comprehensive benchmark for other aspects of spatial reasoning (e.g. topological, distance) and combinations of these; ideally these would be generated programmatically (cf \cite{li2024ijcai}).
\item Consider situations with more than two objects of interest, so that, e.g. compositional reasoning can be tested, and also reasoning about trajectories (cf \cite{weston2016towards}).
\item Look at CD reasoning in a multi-modal setting, an area that has already received some attention, e.g. \citet{10888462}.
\item Investigate LLM performance in reasoning about the cardinal directions between actual geographic entities -- e.g. \citet{Fulman01112024} built a 14 question benchmark to investigate whether LLMs display distortions in estimations of intercardinal directions between cities similarly to the errors made by humans (e.g. estimating whether Reno is east or west of San Diego).
\end{enumerate}
%(e.g. A is east of B, C, C is west of B, what is CD between A and C?); this could then be extended to consider an extended trajectory description, and asking the LLM to compute the CD of the start position with respect to the final location. Also of interest would be to compare LLM performance against human performance.

We would not advocate including ``no yapping'' in the prompt  in future because it may constrain chain of thought in non-reasoning models.

\section*{Data availability}
The data for all experiments (including the data from \cite{cohn_et_al:LIPIcs.COSIT.2024.28}) are available in a GitHub repository at
\url{http://bit.ly/4e069XO}.
%\footnote{
%\url{https://github.com/alan-turing-institute/cosit-2024-evaluating-the-ability-of-llms-to-reason-about-cardinal-directions}}.
\appendix

\section*{Ethical Statement}

There are no ethical issues.

\section*{Acknowledgments}

This work was supported by the Fundamental Research priority area of The Alan Turing Institute,  by  the Economic and Social Research Council (ESRC) under grant ES/W003473/1,  by the EPSRC under grant EP/Z003512/1, and  by the Special Funds of Tongji University for the ``Sino-German Cooperation 2.0 Strategy”.

We also thank Microsoft Research -- Accelerating Foundation Models Research Program for the provision of Azure resources to
test Microsoft-provided models.

%% The file named.bst is a bibliography style file for BibTeX 0.99c
\bibliographystyle{named}
\bibliography{main}

\end{document}